\documentclass[conference]{IEEEtran}
\IEEEoverridecommandlockouts
\usepackage{cite}
\usepackage[linesnumbered,ruled]{algorithm2e}
\usepackage{amsmath,amssymb,amsfonts}
\usepackage{algorithmic}
\usepackage{graphicx}
\usepackage{textcomp}
\usepackage{xcolor}
\usepackage{color}
\usepackage{multirow}
\usepackage{makecell}
\usepackage{booktabs}
\usepackage{url}

\def\BibTeX{{\rm B\kern-.05em{\sc i\kern-.025em b}\kern-.08em
    T\kern-.1667em\lower.7ex\hbox{E}\kern-.125emX}}
\begin{document}

\title{Cooperative coevolutionary hybrid NSGA-II with Linkage Measurement Minimization for Large-scale Multi-objective optimization\\}

\author{\IEEEauthorblockN{Rui Zhong}
\IEEEauthorblockA{\textit{Graduate School of Information Science and Technology} \\
\textit{Hokkaido University}\\
Sapporo, Japan \\
rui.zhong.u5@elms.hokudai.ac.jp}
\and
\IEEEauthorblockN{Masaharu Munetomo}
\IEEEauthorblockA{\textit{Information Initiative Center} \\
\textit{Hokkaido University}\\
Sapporo, Japan \\
munetomo@iic.hokudai.ac.jp}
}
\maketitle

\begin{abstract}
In this paper, we propose a variable grouping method based on cooperative coevolution for large-scale multi-objective problems (LSMOPs), named Linkage Measurement Minimization (LMM). And for the sub-problem optimization stage, a hybrid NSGA-II with a Gaussian sampling operator based on an estimated convergence point is proposed. In the variable grouping stage, according to our previous research, we treat the variable grouping problem as a combinatorial optimization problem, and the linkage measurement function is designed based on linkage identification by the nonlinearity check on real code (LINC-R). We extend this variable grouping method to LSMOPs. In the sub-problem optimization stage, we hypothesize that there is a higher probability of existing better solutions around the Pareto Front (PF). Based on this hypothesis, we estimate a convergence point at every generation of optimization and perform Gaussian sampling around the convergence point. The samples with good objective value will participate in the optimization as elites. Numerical experiments show that our variable grouping method is better than some popular variable grouping methods, and hybrid NSGA-II has broad prospects for multi-objective problem optimization.
\end{abstract}


\section{Introduction}
Multi-objective evolutionary algorithms (MOEAs) have successfully solved various multi-objective problems\cite{Marler:04}. However, the performance of the traditional MOEAs degrades dramatically as the dimension of the problem increases. When the scale of the problem reaches a certain dimension, this type of problem is called large-scale multi-objective optimization problems (LSMOPs). Solving LSMOPs is always tricky, and mainly due to the following aspects: (1). The complexity of optimization problems tends to increase with the increase of dimensions. (2). The search space of large-scale problems increases exponentially with the increase of the dimension, which is known as the curse of dimensionality\cite{Mario:00}. Therefore, a number of generic MOEAs have also been developed for solving LSMOPs since 2013\cite{Antonio:13}. These methods are usually not aimed at a specific type of problem but want to develop a general framework to solve LSMOPs\cite{Tian:21}. There are mainly three strategies currently. (1). Variable grouping method based on Cooperative Coevolution (CC)\cite{Antonio:13}. (2). Methods based on dimensionality reduction or problem transformation\cite{He:19}. (3). Design new search strategies based on MOEAs\cite{Yi:20}. These strategies succeed in solving many-objective optimization problems\cite{Li:15}, constrained multi-objective optimization problems\cite{Fan:17}, and computationally expensive multi-objective optimization problems\cite{Chugh:19}.

This paper mainly adopt CC framework to solve the LSMOPs. Inspired by divide and conquer, the CC framework decomposes the original problem into multiple non-separable sub-problems and applies evolutionary algorithms (EAs) to solve each sub-problem alternately\cite{Potter:94}. This strategy has been a huge success in solving large-scale problems. However, several studies\cite{Omidvar:21A, Omidvar:21B} have shown that the CC framework is sensitive to problem decomposition strategies. In theory, a perfect variable grouping strategy can exponentially reduce the search space without losing optimization accuracy, while a poor grouping strategy will mislead the direction of optimization. Therefore, how to design the decomposition strategy has become a popular research topic.

In addition, unlike large-scale single-objective optimization problems (LSSOPs), in LSMOPs, the same variables in different objective functions may have different interactions. Currently, there are many variable grouping methods were extended from LSSMOPs to group the variables in LSMOPs, such as Fast interdependency identification (FII) for LSMOPs\cite{Li:18}, Differential Grouping (DG) for LSMOPs\cite{Sander:18}, MOEA-DVA\cite{Ma:15}, and other methods. Although these methods show great performance in grouping accuracy, computational cost and local interaction  identification make these algorithms limited. What is more, many studies\cite{Omidvar:21A, Omidvar:21B} in LSSOPs show that high-accuracy grouping is not equivalent to high-performance optimization, and ignoring some weak interactions can accelerate the optimization, especially under the fitness evaluation times (FEs) limitation.

Besides, In order to find the global optimum among the fitness landscape, the heuristic algorithm should be equipped with two major characteristics to ensure finding the global optimum. These two main characteristics are exploration and exploitation\cite{Mirjalili:10}. Exploration is the ability of an algorithm to search whole parts of a problem space whereas exploitation is the convergence ability to the best solution near a good solution. However, the No Free Lunch Theorems (NFLT)\cite{Wolpert:97} has proved that there is no algorithm, which can perform general enough to solve all optimization problems, and a single optimizer cannot balance the exploitation and exploration well in optimization, thus, hybrid evolutionary algorithms and memetic algorithms have become a popular research topic.

In this paper, we propose a novel variable grouping method, named Linkage Measurement Minimization (LMM) and apply hybrid Nondominated Sorting Genetic Algorithm II (hNSGA-II) as the sub-problem optimizer (CC-hNSGA-LMM). In the grouping stage, our proposal allows an automatic decomposition that treats the variable grouping problem as a combinatorial optimization problem, and we design a linkage measure function to evaluate the variable grouping performance. In the sub-problem optimization stage, we introduce a Gaussian sampling operator based on a estimated convergence point combined with NSGA-II to balance the exploitation and exploration in optimization. Specifically, the main contributions of this paper are as follows.

(1) For variable grouping, we propose a novel automatic variable grouping strategy named LLM and explain the mathmatical mechanism of LLM in detial, which reveals the relationship between our proposal and LINC-R. 

(2) For optimizer, we first hypothesize that the individuals around the PF have better fitness, and the participation of elite individuals in optimization can accelerate the convergence. Based on this hypothesis, we estimate a convergence point at every generation of optimization and apply the Gaussian sampling as a local search operator to exploit the potential individuals around the estimated convergence point.

(3) Theoretical analysis and numerical experiments show that our proposal has broad prospects in solving LSMOPs and can solve larger-scale and many-objective optimization problems through simple extension.

The rest of the paper is organized as follows, Section II covers preliminaries and a brief review of grouping methods. Section III introduces our proposal in detail. Section IV shows the experiments and analysis. Section V discusses about the direction of our research in the future, Section VI concludes the paper and shows future directions.

\section{Preliminaries and related works}
Many grouping methods have been proposed based on CC for LSSOPs in recent years\cite{Omidvar:21A, Omidvar:21B}, and meanwhile, some works have been expanded to LSMOPs\cite{Li:18,Tian:21}. In this section, we firstly introduce some preliminaries including the concept of multi-objective problems, separability of functions, estimation of a convergence point, and NSGA-II. Then, we will provide a brief review of grouping methods both in LSSOPs and LSMOPs.

\subsection{Preliminaries}

\subsubsection{Multi-objective Problems}
Without loss of generality, a Multi-objective Problem (MOP) can be mathematically defined as Eq (\ref{eq:1}):
\begin{equation}
	\begin{aligned}
		\label{eq:1}
		\min \quad F({\rm x})=(f_{1}({\rm x}),f_{2}({\rm x}),...,f_{M}({\rm x})), \\
		s.t. \quad {\rm x} \in \Omega,
	\end{aligned}
\end{equation}
Where $f:\Omega \to \Lambda \subseteq \mathbb{R}^{M}$ consists of M objectives, $\Lambda$ is the objective space, $\Omega \subseteq \mathbb{R}^{D}$ is the decision space, and ${\rm x}=(x_{1},x_{2},...,x_{D}) \in \Omega$ is a solution consisting of $D$ decision variables. The dominance relation between two solutions can be defined as $(\forall i \in {1,...,M},f_{i}({\rm x}) \leqslant f_{i}({\rm y}) \wedge (\exists j \in {1,...,M}, f_{i}({\rm x}) \leq f_{i}({\rm y}))$. If this formula is satisfied, we can say that x dominates y. A Pareto optimal solution is a solution that is not dominated by any solution in $\Omega$

\subsubsection{Separability of functions}
The concept of separability is derived from a biological concept\cite{Tezuka:04}. if a feature at the phenotype level is determined by two or more genes, then we say that these genes have interaction. Then we extend the concept to optimization problems. When $f(x_{1},x_{2},...,x_{n})=\sum_{i=1}^{m}f(x_{i_{1}},...,x_{i_{k}})$, We call $f(x)$ a partially separable function, and the variables in each sub-problem are called linkage sets\cite{Hillol:96}. The explicit or implicit interactions exist between every pair of variables in a linkage set, and we call these variables non-separable variables, and there is no interaction between variables in different linkage sets, and these variables are called separable variables. There are two extreme cases of the interaction, when there is no interaction between all variables, $f(x_{1},x_{2},...,x_{n})=\sum_{i=1}^{n}f(x_i)$, we call $f(x)$ is a fully separable function. On the contrary, when every pair of variables exist explicit or implicit interaction, we call $f(x)$ a fully non-separable function. Based on CC framework, we want to develop a grouping method which can find the interactions and divide the variables into sub-problems properly.

\subsubsection{Estimation of a convergence point}
\begin{figure}[htb]
	\centering
	\includegraphics[width=9cm]{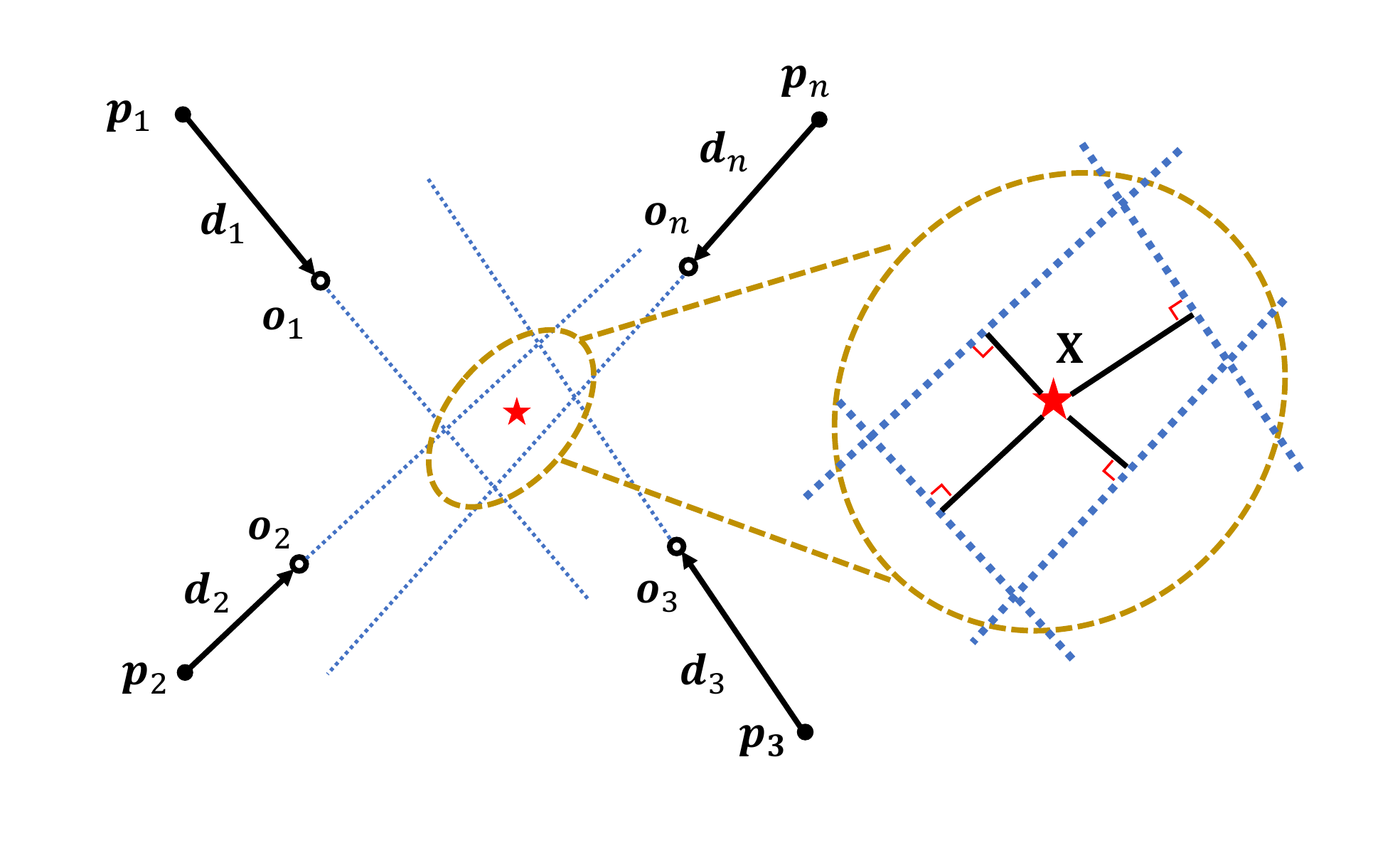}
	\caption{Moving vector $\boldsymbol{d}_i (=\boldsymbol{o}_i-\boldsymbol{p}_i)$ is calculated from a parent (worse) individual $\boldsymbol{p}_i$ and its offspring (better) $\boldsymbol{o}_i$. The $\star$ is the estimated convergence point.}
	\label{fig:10}
\end{figure}

Estimation of a convergence point was first proposed by Murata. The paper\cite{Noboru:15} hypothesize that: in a population-based optimization algorithm, all individuals are moving towards the global optimum. Fig.\ref{fig:10} shows how the estimation of a convergence point works. Although the estimated point is not exactly on the global optimum in practice due to incorrect directions or inaccurate population movements. However, it has higher possibility that the estimated point is close to the global optimum.

Let us derive how to estimate the convergence point mathematically. First, we define parent (worse) individual $\boldsymbol{p}_i$, offspring (better) $\boldsymbol{o}_i$, and moving vector $\boldsymbol{d}_i$ as describe in Fig \ref{fig:10}. The unit direction vector of $\boldsymbol{d}_i$ is given as $\boldsymbol{d}_{0i}=\frac{\boldsymbol{d}_i}{||\boldsymbol{d}_i||}$,i.e., $\boldsymbol{d}_{0i}^{\rm T}\boldsymbol{d}_{0i}=1$. $\mathbf{X}$ denotes the estimated convergence point, and $\boldsymbol{p}_i+t_i\boldsymbol{d}_i$ represents the expansion from parent individual $\boldsymbol{p}_i$ with the direction $\boldsymbol{d}_i$. $L(\mathbf{X},{t_i})$ in Eq (\ref{eq:19}) becomes the minimum.
\begin{equation}
	\begin{aligned}
		\label{eq:19}
		\min (L(\mathbf{X},{t_i})) = \min (\sum_{i=1}^{n}||\boldsymbol{p}_i+t_i\boldsymbol{d}_i-\mathbf{X}||^2)
	\end{aligned}
\end{equation}

As the minimum line segment from the convergence point $\mathbf{X}$ to the expansion line segments is the orthogonal projection from $\mathbf{X}$, we can apply the Eq (\ref{eq:20}) into Eq (\ref{eq:19}) to remote $t_i$.
\begin{equation}
	\begin{aligned}
		\label{eq:20}
		\boldsymbol{d}_i^T(\boldsymbol{p}_i+t_i\boldsymbol{d}_i-\mathbf{X})=0 \ ({\rm orthogonal \ condition})
	\end{aligned}
\end{equation}

Finally, the convergence point $\mathbf{X}$ can be calculated by Eq (\ref{eq:21}). See detail expansion of equations in paper \cite{Noboru:15}.
\begin{equation}
	\begin{aligned}
		\label{eq:21}
		\widehat{\mathbf{X}}=\left\lbrace \sum_{i=1}^{n}(\boldsymbol{I}_d-\boldsymbol{d}_{0i}\boldsymbol{d}_{0i}^{\rm T}) \right\rbrace ^{-1}\left\lbrace \sum_{i=1}^{n}(\boldsymbol{I}_d-\boldsymbol{d}_{0i}\boldsymbol{d}_{0i}^{\rm T})\boldsymbol{p}_i\right\rbrace
	\end{aligned}
\end{equation}

\subsubsection{NSGA-II}
Non-dominated Sorting Genetic Algorithm (NSGA-II) is proposed by Deb in 2002 to solve MOPs\cite{Deb:02}. Now, NSGA-II is one of the most popular MOEAs with three special characteristics, the fast non-dominated sorting approach, the fast crowded distance estimation procedure, and the simple crowded comparison operator. Deb et al. simulated several test problems from previous studies using the NSGA-II optimization technique and showed exciting results. The pseudocode of NSGA-II is described in Algorithm \ref{alg:1}

\begin{algorithm}
	\label{alg:1}
	\DontPrintSemicolon
	\SetAlgoLined
	\KwIn {${\rm Population}:P$}
	\KwOut {${\rm Pareto \ Front}:PF$}
	\SetKwFunction{FNSGA}{\textbf{NSGA-II}}
	\SetKwProg{Fn}{Function}{:}{}
	\Fn{\FNSGA{$s$}}{
		$t \gets 0$\;
		$Q_{t} \gets \emptyset $\;
		$PF \gets $\textbf{nonDominated}($P_{t}$)\;
		\While{{\rm not stop criterion}}{
			$R_{t} \gets P_{t} \cup Q_{t}$\;
			$\mathcal{F} \gets $ \textbf{fastNonDominateSorting}($R_{t}$)\;
			$P_{t+1} \gets \emptyset$\;
			$i \gets 1 $\;
			\While{$|P_{t+1}|+|\mathcal{F}_{i}| \leqslant s$}{
				$\mathcal{C}_{i} \gets $ \textbf{crowdingDistanceAssigment}($\mathcal{F}_{i}$)\;
				$P_{t+1} \gets P_{t} \cup \mathcal{F}_{i}$\;
				$i \gets i+1 $\;
			}
			$\mathcal{F}_{i} \gets $\textbf{Sort}($\mathcal{F}_{i},\mathcal{C}_{i}$)\;
			$P_{t+1} \gets P_{t+1} \cup \mathcal{F}_{i}[1:(N-|P_{t+1}|)] \rhd $ fill $P_{t+1}$ with the $N-|P_{t+1}|$ less crowded individuals of $\mathcal{F}_{i}$ \;
			$Q_{t+1} \gets$ \textbf{Selection}($P_{t+1}, s$)\;
			$Q_{t+1} \gets$ \textbf{Crossover}($Q_{t+1}$)\;
			$Q_{t+1} \gets$ \textbf{Mutation}($Q_{t+1}$)\;
			$t \gets t+1 $\;
			$PF \gets $ \textbf{nonDominated}($A \cup Q_{t}$)\;
		}

		$\textbf{return} \ PF$
	}
	\caption{NSGA-II}
\end{algorithm}

\subsection{A brief review of grouping methods}
Based on the idea of divide and conquer, the CC framework was proposed in 1994\cite{Potter:94} and has been popularized in large-scale optimization. CC requires decomposing the problem into a set of low-dimensional subproblems, each of which is optimized separately. Since the candidate solutions of each sub-component cannot form a complete solution, representative solutions of other sub-components are required to form a complete solution for evaluation. These representative solutions are known as the context vector\cite{Van:04}. The context vector is updated iteratively and acts as the context in which the cooperation occurs. In this part of the section, we briefly introduce the grouping principle on LSSOPs, which can be classified into three major groups: automatic, semi-automatic, and $m \times k$-strategy.

\subsubsection{Automatic}
In the automatic grouping methods, the formation of sub-components completely depends on the logic of the algorithm. The representative algorithms include LINC-R\cite{Tezuka:04} and LIMD\cite{Munetomo:99}, and are extended to DG\cite{Omidvar:14}, DG2\cite{Omidvar:17}, ERDG\cite{Yang:21}, and CCVIL\cite{Chen:10}, etc. respectively. These methods are mainly designed based on perturbations. Eq (\ref{eq:3}) defines a sample $s$ perturbs $\delta$ in $i^{th}$-D and $j^{th}$-D and both in $i^{th}$-D and $j^{th}$-D.
\begin{equation}
	\label{eq:3}
	\begin{aligned}
		s=(x_{1},x_{2},...,x_{n}) \\
		s_{i}=(x_{1},...,x_{i}+\delta,...,x_{n}) \\
		s_{j}=(x_{1},...,x_{j}+\delta,...,x_{n}) \\
		s_{ij}=(x_{1},...,x_{i}+\delta,...,x_{j}+\delta,...,x_{n}) \\
	\end{aligned}
\end{equation}

Eq (\ref{eq:4}) is employed by LINC-R to identify the interaction between $x_i$ and $x_j$.
\begin{equation}
	\label{eq:4}
	\begin{aligned}
		\exists s \in Pop: \\
		\Delta_{i} = f(s_{i}) - f(s) \\
		\Delta_{j} = f(s_{j}) - f(s) \\
		\Delta_{ij} = f(s_{ij}) - f(s) \\
		if |\Delta_{ij} - (\Delta_{i}+\Delta_{j})| > \epsilon \\
		then \  x_{i} \  and \  x_{j} \  are \  nonseparable
	\end{aligned}
\end{equation}
$\epsilon$ is the allowable error. LIMD applied Eq (\ref{eq:5}) to detect the interaction between $x_i$ and $x_j$.
\begin{equation}
	\label{eq:5}
	\begin{aligned}
		\exists s \in Pop: \\
		if \ \neg(f(s) < f(s_i) < f(s_{ij}) \ and \ f(s) < f(s_j) < f(s_{ij}) \\
		or \ f(s) > f(s_i) > f(s_{ij}) \ and \ f(s) > f(s_j) > f(s_{ij})) \\
		then \  x_{i} \  and \  x_{j} \  are \  nonseparable
	\end{aligned}
\end{equation}
When $x_{i}$ and $x_{j}$ not satisfy the simultaneous increase or decrease on at least one individual of population, LIMD identifies $x_{i}$ and $x_{j}$ as nonseparable variables. Notice that we cannot apply LINC-R and LIMD to the whole fitness landscape in practice, and only finite individuals $s$ are checked. In other words, if the condition both in LINC-R and LIMD are not satisfied, then $x_{i}$ and $x_{j}$ are identified as separable variables.

\subsubsection{Semi-automatic}
Semi-automatic methods decompose the problem depending on the both algorithm logic and parameters specified by users. Some studies apply statistical detection methods to form the sub-components by a threshold or a set of intervals defined on correlation coefficients with the participation of users, such as AVP2\cite{Singh:10} and 4CDE\cite{Rojas:11}. The fuzzy c-mean algorithm proposal by Fan et al\cite{Fan:14}. forms the sub-components with the number of components. This kind of algorithm often consumes fewer FEs to detect the interactions, which is an advantage compared with automatic methods. 

\subsubsection{$m \times k$-strategy}
This kind of algorithm requires fewer FEs to form the sub-components than automatic methods and semi-automatic methods. The number and the size of each component are necessary hyperparameters decided by users. And the representative detection principles include Random Grouping\cite{Yang:08,Nabi:10}, Delta Grouping\cite{Omidvar:10}, Fitness Difference Partitioning\cite{Sayed:15, Dai:16}, and so on. 

CC framework is a well-studied technique for LSSOPs, and in LSMOPs, due to the multiple objective functions, and the variable interactions in different objective functions are often different, so the grouping methods developed for LSSOPs always need to be modified for LSMOPs. Currently, the popular grouping strategies on LSMOPs include Random Grouping\cite{Antonio:13}, Differential Grouping\cite{Sander:18}, and Variables Analysis\cite{Ma:15}. In paper \cite{Song:16}, dynamic Random Grouping is applied. The variables are regrouped after each generation of optimization. As same as the Random Grouping in LSSOPs, the probability of two interacting variables being divided in the same sub-problem is quite high in multiple trial experiments. In DG for LSMOPs, the paper\cite{Li:18} applies Eq (\ref{eq:4}) to all objective functions, and when $x_{i}$ and $x_{j}$ satisfy Eq (\ref{eq:4}) in all objective functions, they are identified as separable variables. Both Random Grouping and DG were originally proposed to solve LSSOPs, and these methods focus on partitioning decision variables into sub-problems correctly while ignoring the population diversity in the objective space. Therefore, the variable grouping methods based on Random Grouping or DG can easily find some local or global optimal solutions but may not be able to diversify the population along the whole Pareto front. Variable Analysis (VA) is a grouping method for LSMOPs. MOEA/DVA\cite{Ma:15} perturbs several times on a random sample. If all the perturbed samples are non-dominated with each other, this variable is considered a position variable, and if each perturbed sample is dominated or dominating the rest samples, this variable is regarded as a distance variable, otherwise, it is regarded as a mixed variable. MOEA/DVA optimizes the different types of variables in order and allocates different computational resources. Numerical experiments show that MOEA/DVA significantly outperforms many other MOEAs on LSMOPs in benchmarks.

\section{CC-hNSGA-LMM}
\begin{figure*}[htb]
	\centering
	\includegraphics[width=17cm]{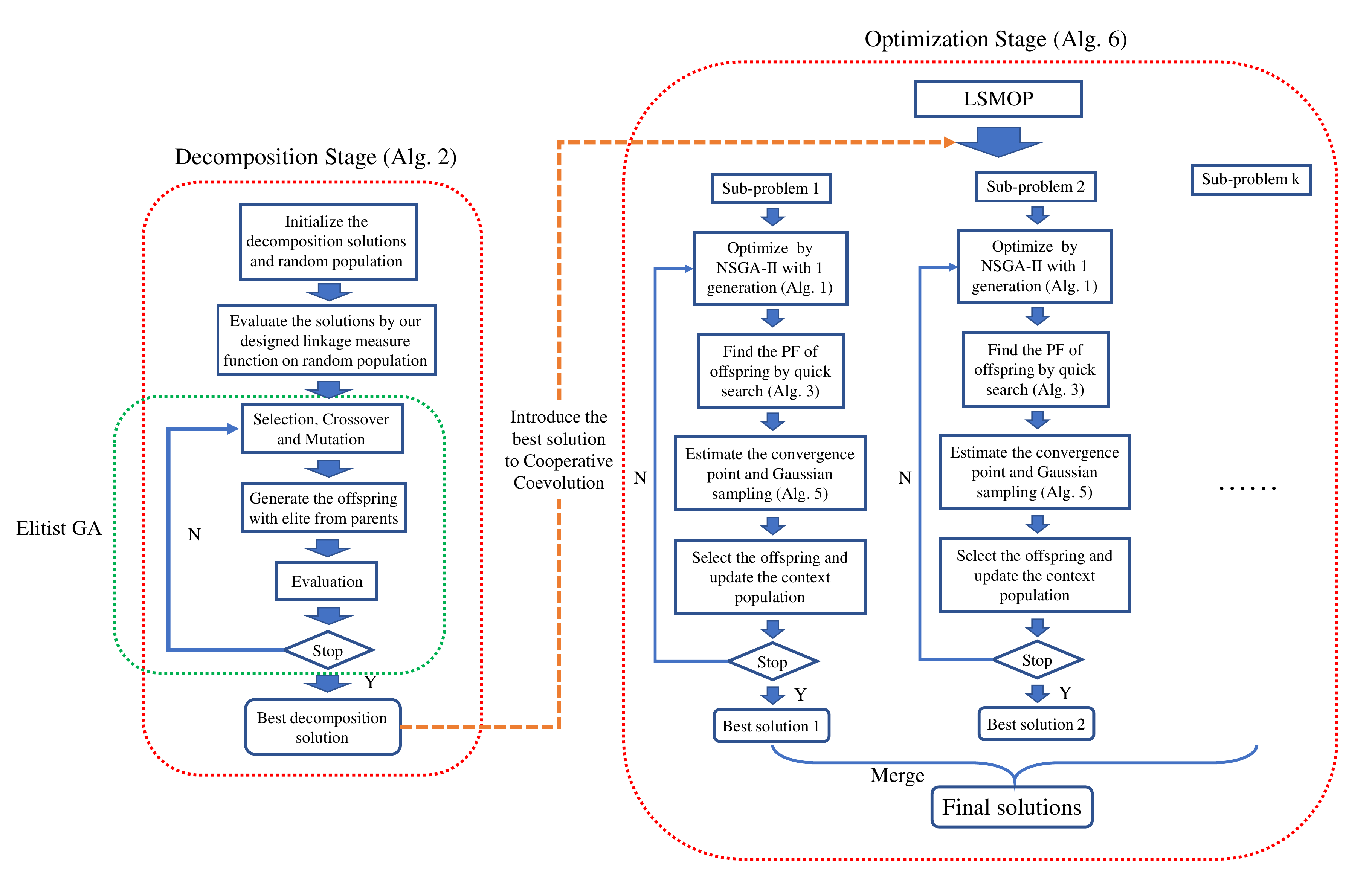}
	\caption{The flowchart of CC-hNSGA-LMM}
	\label{fig:1}
\end{figure*}

In this section, we will introduce the details of our proposal and the techniques. The flowchart of our proposal is shown in Figure \ref{fig:1}.

Our proposal includes the decomposition stage and the optimization stage. Next, we will introduce the flow of our proposal in detail.

In decomposition stage, our previous research\cite{Zhong:22} regards the decomposition problem as an optimization problem and designed the linkage measurement function based on LINC-R. Next, we will give a simple mathematical explanation of our proposal.

In LINC-R, we can rewrite Eq (\ref{eq:4}) to Eq (\ref{eq:7})
\begin{equation}
	\label{eq:7}
	\begin{aligned}
		\exists s \in Pop: \\
		if |(f(s_{ij})-f(s_{i})) - (f(s_{j})-f(s))| > \epsilon \\
		then \  x_{i} \  and \  x_{j} \  are \  nonseparable
	\end{aligned}
\end{equation}
And we notice that LINC-R can be understood as the additive form of vector. Eq (\ref{eq:7}) can also be written to Eq (\ref{eq:8})
\begin{equation}
	\label{eq:8}
	\begin{aligned}
		\exists s \in Pop: \\
		if |(f(s_{ij})-f(s)) - ((f(s_{i})-f(s))+(f(s_{j})-f(s)))| > \epsilon \\
		then \  x_{i} \  and \  x_{j} \  are \  nonseparable
	\end{aligned}
\end{equation}

\begin{figure}[htb]
	\centering
	\includegraphics[width=9cm]{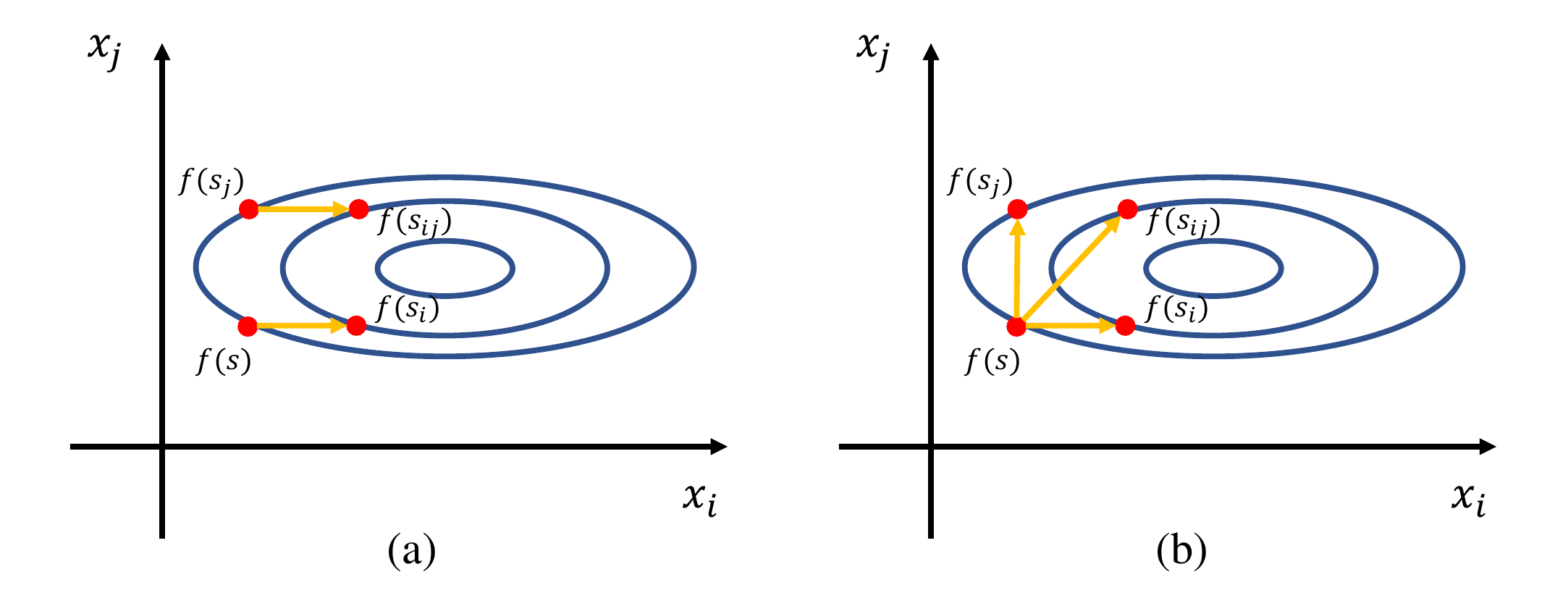}
	\caption{(a).The original LINC-R works on the separable variables. (b).The variant LINC-R works on the separable variables.}
	\label{fig:2}
\end{figure}

Figure \ref{fig:2} shows how original LINC-R and variant LINC-R work on separable variables.

\begin{figure}[htb]
	\centering
	\includegraphics[width=9cm]{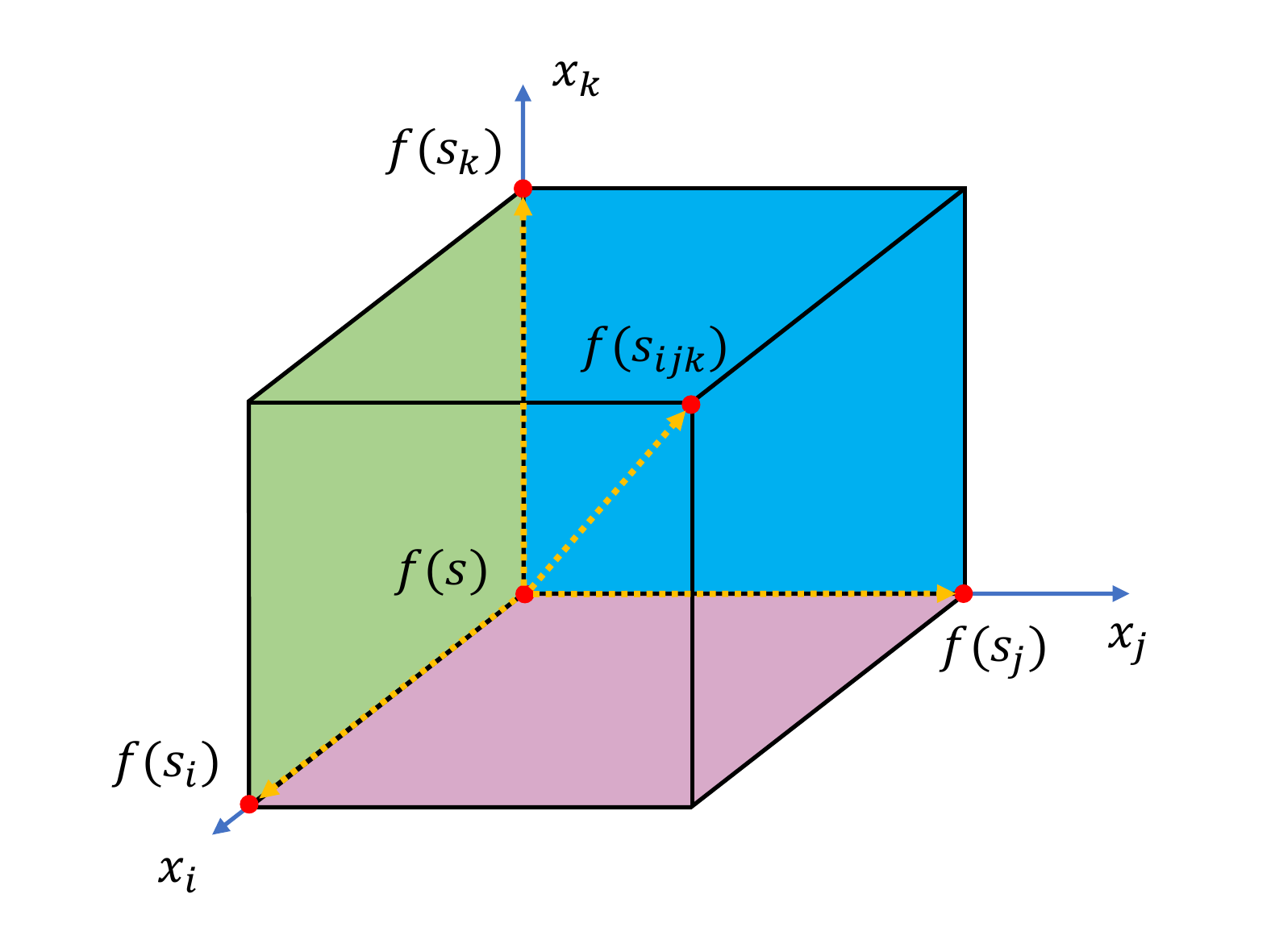}
	\caption{The variant LINC-R works on 3-D space}
	\label{fig:3}
\end{figure}

Next, we derive the variant LINC-R to $3$-D or higher dimensions. In $3$-D space, the schematic diagram is shown in Figure \ref{fig:3}.
Similarly, we define the fitness difference in $3$-D space in Eq (\ref{eq:9}) 
\begin{equation}
	\label{eq:9}
	\begin{aligned}
		s \in Pop: \\
		\Delta f_{i} = f(s_{i}) - f(s) \\
		\Delta f_{j} = f(s_{j}) - f(s) \\
		\Delta f_{k} = f(s_{k}) - f(s) \\
		\Delta f_{ijk} = f(s_{ijk}) - f(s) 
	\end{aligned}
\end{equation}
And Eq (\ref{eq:10}) defines the variant LINC-R in $3$-D space
\begin{equation}
	\label{eq:10}
	\begin{aligned}
		\exists s \in Pop: \\
		if |\Delta_{ijk} - (\Delta_{i}+\Delta_{j}+\Delta_{k})| > \epsilon \\
		then \ interaction(s) \ exist \ in \ x_i, x_j, x_k
	\end{aligned}
\end{equation}

Therefore, we can reasonably infer the variant LINC-R in $n$-D space on Eq (\ref{eq:11}).
\begin{equation}
	\label{eq:11}
	\begin{aligned}
		\exists s \in Pop: \\
		if |\Delta_{1,2,...,n} - (\Delta_{1}+\Delta_{2}+...+\Delta_{n})| > \epsilon \\
		then \ interaction(s) \ exist \ in \  x_{1},x_{2},...,x_{n} 
	\end{aligned}
\end{equation}

Notice that we only detect the interactions based on the finite individuals, which means when $|\Delta_{1,2,...,n} - (\Delta_{1}+\Delta_{2}+...+\Delta_{n})| > \epsilon$ is not satisfied at least once in all individuals, then we consider this function is a fully separable function by default. Although this strategy is limited especially for trap functions, it is impossible to check the interactions on the whole fitness landscape, and in LSSOPs, perturbation-based methods often identify the interactions based on 1 sample such as DG, DG2, etc. due to the FEs. Thus, Eq (\ref{eq:22}) is approximately correct in LSSOPs.
\begin{equation}
	\label{eq:22}
	\begin{aligned}
		\forall s \in Pop: \\
		if |\Delta_{1,2,...,n} - (\Delta_{1}+\Delta_{2}+...+\Delta_{n})| < \epsilon \\
		then  x_{1},x_{2},...,x_{n} are \ separable
	\end{aligned}
\end{equation}

However, when Eq (\ref{eq:22}) is not satisfied, we only know that interactions exist in some variable pairs, but we cannot know the interactions exist in which pairs, so we can actively to detect the interactions between variables. Taking the $3$-D space as an example,
\begin{equation}
	\label{eq:12}
	\begin{aligned}
		\forall s \in Pop: \\
		if |\Delta_{ijk} - (\Delta_{i}+\Delta_{j}+\Delta_{k})| > \epsilon \\ and \ |\Delta_{ijk} - (\Delta_{ij}+\Delta_{k})| < \epsilon \\ 
		then \  x_{i},x_{j} \ are \ non-separbale \\
		and \ x_{k} \ is \ separable \ from \ x_{i},x_{j}
	\end{aligned}
\end{equation}
Therefore, Our target is to apply the heuristic algorithm to find the interactions between all variables as much as possible. According to the above explanation, in the $n$-D problem, the linkage measurement function in our proposal is designed as Eq (\ref{eq:13})
\begin{equation}
	\label{eq:13}
	\begin{aligned}
		\min ((\Delta_{1,2,...,n}-\sum^{m}(\Delta_{i,...,k}))^2) 
	\end{aligned}
\end{equation}
$m$ is the number of sub-problems. Eq (\ref{eq:13}) is the original linkage measurement function of our proposal\cite{Zhong:22}. Meanwhile, our further research notice that this linkage measurement function often contains multiple optima especially in separable functions and partially separable function. Therefore, we can attach a reasonable penalty to lead the direction of optimization. Eq (\ref{eq:15}) defines a linkage measurement function with a penalty term.
\begin{equation}
	\label{eq:15}
	\begin{aligned}
		\min( \frac{|\Delta_{1,2,...,n}-\sum^{m}_{i,j,...}\Delta_{i,j,..}|}{num \ of \ group})
	\end{aligned}
\end{equation}
Then, we extend Eq (\ref{eq:15}) to LSMOPs with multiple samples. Eq (\ref{eq:16}) is our linkage measurement function in this paper.
\begin{equation}
	\label{eq:16}
	\begin{aligned}
		\min(\sum_{s \in Pop}\sum_{j=1}^{M}w_j \frac{|\Delta_{1,2,...,n}-\sum^{m}_{i,j,...}\Delta_{i,j,..}|}{num \ of \ group}),	\ \sum_{j=1}^{M}w_j = 1
	\end{aligned}
\end{equation}
$w_j$ is the weight of the $j^{th}$ objective function, and $M$ is the number of objective functions in LSMOPs. We apply averaging weight in Eq (\ref{eq:16}).

To optimize this linkage measurement function, we also apply Elitist Genetic Algorithm (EGA)\cite{De:75}. The pseudocode of decomposition stage is shown in Algorithm \ref{alg:2}

\begin{algorithm}
	\label{alg:2}
	\DontPrintSemicolon
	\SetAlgoLined
	\KwIn {${\rm Dimension}:D;{\rm Population \ size}:s_1;{\rm Sample \ size}:s_2;{\rm Gene \ length}:L;{\rm Generation}:T$}
	\KwOut {${\rm The \ best \ decomposition}:E$}
	\SetKwFunction{FLMM}{\textbf{LMM}}
	\SetKwProg{Fn}{Function}{:}{}
	\Fn{\FLMM{$D, s_1, s_2, L, T$}}{
		$t \gets 0$\;
		$\blacktriangleright$ (Initialization) \;
		\For{$i=0 \ to \ s_1$}{
			\For{$j=0 \ to \ D$}{
				$n \gets \textbf{randint}(0, 2^{L-1}-1)$ \;
				$P_{i,t}^n \gets j$ \;
			}
		}
		$E \gets \textbf{initialCC}(D)$ \;
		$S \gets \textbf{randSamples}(s_2)$\;
		$FE \gets $ \textbf{Evaluate}($E, S$) \;
		$\blacktriangleright$ (Fully separable function) \;
		\If{$FE < 0.01$}{
			$\textbf{return} \ E$
		}
		$F_{t} \gets $ \textbf{Evaluate}($P_{t}^n, S$) \;
		$E \gets $ \textbf{bestIndividual}($P_{t}^n, E$) \;
		$\blacktriangleright$ (Optimization) \;
		\While{{\rm not stop criterion}}{
			$P_{t+1} \gets$ \textbf{Selection}($P_{t},F_{t},M$) \;
			$P_{t+1} \gets$ \textbf{Crossover}($P_{t+1}$) \;
			$P_{t+1} \gets$ \textbf{Mutation}($P_{t+1}$) \;
			$F_{t+1} \gets $ \textbf{Evaluate}($P_{t+1}, S$) \;
			$P_{t+1} \gets $ \textbf{Replace}($P_{t+1}, E$) \;
			$E \gets $ \textbf{bestIndividual}($P_{t+1}, E$) \;
			$t \gets t+1 $ \;
		}
		$\textbf{return} \ E$
	}
	\caption{Decomposition Stage}
\end{algorithm}
 
The elitist reservation strategy directly replicates the best individual without crossover, mutation, and selection to the next generation. This strategy can prevent the optimal individual from destroying the superior gene and chromosome structure during crossover mutation.

In optimization stage, first we hypothesize that there is much possible for better individuals around the PF individuals. This hypothesis is extended from single objective optimization problems\cite{Noboru:15}. Based on this hypothesis, we find the PF in every generation of optimization by quick search. The pseudocode of quick search is shown in Algorithm \ref{alg:3}.

\begin{algorithm}
	\label{alg:3}
	\DontPrintSemicolon
	\SetAlgoLined
	\KwIn {${\rm Population}:P;{\rm Objective \ value}:O;{\rm num \ of \ objective \ function}:M$}
	\KwOut {${\rm Pareto \ Front}: PF$}
	\SetKwFunction{FQS}{\textbf{QS}}
	\SetKwProg{Fn}{Function}{:}{}
	\Fn{\FQS{$P, O, M$}}{
		$s \gets \textbf{size}(P)$ \;
		$\blacktriangleright$ (All solutions are PF in initialization) \;
		$R \gets [1] * s$\;
		\For{$i=0 \ to \ s$}{
			\If{$R_i != 1$}{
				\textbf{continue}\;
			} 
			\For{$j=i+1 \ to \ s$}{
				\If{$R_j != 1$}{
					\textbf{continue}\;
				} 
				$D \gets \textbf{Dominate}(O_i, O_j, M)$ \;
				\If{$D == 1$}{
					$R_i \gets 0$\;
					\textbf{break}\;
				} \ElseIf{$D == -1$} {
					$R_j \gets 0$\;
				} \Else{
					\textbf{continue}\;
				}
			}
		}
		$t \gets 0$\;
		\For{$i=0 \ to \ s$}{
			\If{$R_i == 1$}{
				$PF_t \gets P_i$\;
				$t \gets t+1$\;
			}
		}
		$\textbf{return} \ PF$
	}
	\caption{Quick Search}
\end{algorithm}
And the function Dominate is realized in Algorithm \ref{alg:4}.
\begin{algorithm}
	\label{alg:4}
	\DontPrintSemicolon
	\SetAlgoLined
	\KwIn {${\rm Objective \ value \ of \ } i^{th} {\rm \ individual}:O_i; {\rm Objective \ value \ of \ } j^{th} {\rm \ individual}:O_j; {\rm num \ of \ objective \ function}:M$}
	\KwOut {${\rm Domination} \ D: 0 \ ({\rm No \ domination});$ $-1 \ (i^{th}{ \ \rm Dominates\ }j^{th}); $ $1 \ (j^{th}{ \ \rm Dominates\ }i^{th}) $}
	\SetKwFunction{FD}{\textbf{Dominate}}
	\SetKwProg{Fn}{Function}{:}{}
	\Fn{\FD{$O_i, O_j, M$}}{
		$IJ \gets 0$ \;
		$JI \gets 0$ \;
		$D \gets 0$ \;
		\For{$k=0 \ to \ M$}{
			\If{$O_i^k >= O_j^k$}{
				$JI \gets JI+1$ \;
			}
			\If{$O_j^k >= O_i^k$}{
				$IJ \gets IJ+1$ \;
			}
		}
		\If{$IJ == M$ \textbf{and} $JI != M$}{
			$D \gets -1$\;
		} \ElseIf{$IJ != M$ \textbf{and} $JI == M$} {
			$D \gets \ 1$\;
		} \Else{
			$D \gets \ 0$\;
		}
		$\textbf{return}  \ D$\;
	}
	\caption{Domination identification}
\end{algorithm}

\begin{figure}[htb]
	\centering
	\includegraphics[width=9cm]{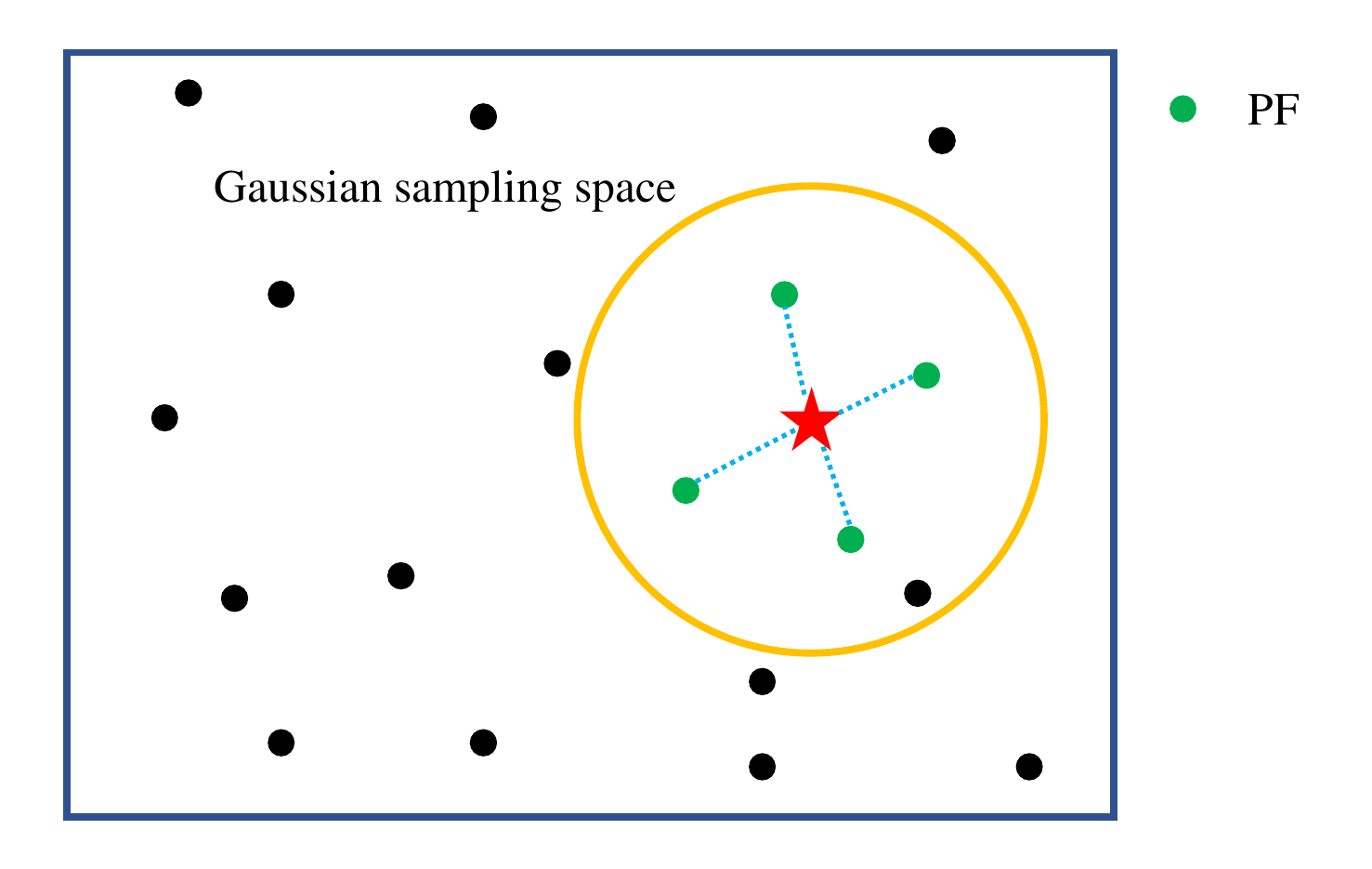}
	\caption{The averaging strategy to estimate the convergence point approximately.}
	\label{fig:4}
\end{figure}

The target of our quick search is only to find the whole PF in the current population, thus the fast non-dominated sorting in this situation is wasteful and unnecessary. From Algorithm \ref{alg:3}, the worst and best time complexity of quick search is $O(MN^2)$ and $O(MN)$ respectively. $M$ is the number of objective functions, and $N$ is the population size. After the PF is found, we estimate the convergence point with averaging strategy and apply the Gaussian sampling with the mean of the estimated convergence point. Figure \ref{fig:4} and Algorithm \ref{alg:5} shows how our proposal works. 

\begin{algorithm}
	\label{alg:5}
	\DontPrintSemicolon
	\SetAlgoLined
	\KwIn {${\rm Dimension}:D;{\rm Pareto \ Front}:PF; {\rm Sampling \ size}:s$}
	\KwOut {${\rm Samples}:S$}
	\SetKwFunction{FEGS}{\textbf{EGS}}
	\SetKwProg{Fn}{Function}{:}{}
	\Fn{\FEGS{$D, PF, s$}}{
		$p \gets [0] * D$ \;
		$n \gets \textbf{size}(PF)$ \;
		$\blacktriangleright$ (Estimate a convergence point with averaging) \;
		\For{$i=0 \ to \ D$}{
			\For{$j=0 \ to \ n$}{
				$p_i \gets p_i + PF_{i,j}$ \;
			}
			$p_i \gets p_i / n$ \;
		}
		$S \gets \textbf{GaussianSampling}(p, s)$ \;
		$\textbf{return}  \ S$\;
	}
	\caption{Estiamte a convergence point and apply Gaussian sampling}
\end{algorithm}

Finally, the whole process of our proposal in optimization stage is described on Algorithm \ref{alg:6}

\begin{algorithm}
	\label{alg:6}
	\DontPrintSemicolon
	\SetAlgoLined
	\KwIn {${\rm Dimension}:D;{\rm LSMOP}:MP;{\rm Population \ size}:s; {\rm Max \ iteration}:G;{\rm Decomposition \ solution}:E; {\rm num \ of \ objective \ function}:M$}
	\KwOut {${\rm Pareto \ Front}:PF$}
	\SetKwFunction{FOPT}{\textbf{OPT}}
	\SetKwProg{Fn}{Function}{:}{}
	\Fn{\FOPT{$D, MP, s, G, E$}}{
		$Ps \gets \textbf{Decomposition}(MP, E)$ \;
		$n \gets \textbf{size}(Ps)$ \;
		$CP \gets \emptyset$   // context population \;
		$PF \gets \emptyset$ \;
		\For{$i=0 \ to \ n$}{
			$P_i \gets \textbf{randPop}(s)$ \;
			\For{$j=0 \ to \ G$}{
				$Q_{i,j} \gets \textbf{NSGA-II}(P_i)$ // Population before non-Dominate sorting \;
				$O_{i,j} \gets \textbf{Evaluate}(Q_{i,j}, CP)$ \;
				$PF_{i,j} \gets \textbf{QS}(Q_{i,j}, O_{i,j}, M)$ \;
				$S \gets \textbf{EGS}(D, PF_{i,j}, s/10)$ \;
				$OS_{i,j} \gets \textbf{Evaluate}(S, CP)$ \;
				$Q_{i,j} \gets Q_{i,j} \cup S$ \;
				$P_{i+1} \gets \textbf{fastNonDominateSorting}(Q_{i,j})$ \;
				$CP \gets \textbf{Update}(P_{i+1})$ \;
			}
			$PF \gets \textbf{Update}(CP)$ \;
		}
		$\textbf{return}  \ PF$
	}
	\caption{Optimization Stage}
\end{algorithm}

\section{Experiment results and analysis}
In this Section, we ran many experiments to evaluate our proposal, CC-hNSGA-LMM. In Section IV-A, we introduce the experiment settings, including benchmark functions, comparing methods, parameters of algorithms, and the performance indicators. In Section IV-B, we provide the experiment results. Finally, we analyze our proposal both in decomposition stage and optimization stage in Section IV-C.

\subsection{Experiment Settings}

\subsubsection{Benchmark functions}
We conduct our experiments on benchmark functions up to $500$-D and $1000$-D. The details of benchmark functions are shown in Table \ref{tbl:1}

\begin{table}[tbh]
	\scriptsize
	\centering
	\caption{The benchmark functions of our experiment}
	\label{tbl:1}
	\begin{tabular}{cccc}
		\toprule
		Test Suite & Benchmark & Feature of PF & Separability \\
		\midrule
		\multirow{6}*{ZDT\cite{Zitzler:00}}& ZDT1 & Convex  & Separable	\\ 
		~ & ZDT2 & Concave  & Separable	\\ 
		~ & ZDT3 & Convex, disconnected  & Separable	\\ 
		~ & ZDT4 & Convex  & Separable	\\
		~ & ZDT5 & Convex  & Separable	\\
		~ & ZDT6 & Concave & Separable	\\ 
		\midrule
		\multirow{7}*{DTLZ\cite{Deb:05}} & DTLZ1 & Linear  & Separable	\\ 
		~ & DTLZ2 & Concave  & Separable	\\ 
		~ & DTLZ3 & Concave  & Separable	\\ 
		~ & DTLZ4 & Concave  & Separable	\\ 
		~ & DTLZ5 & Concave, degenerate  & Separable	\\
		~ & DTLZ6 & Concave, degenerate  & Separable	\\ 
		~ & DTLZ7 & Disconnected  & Separable	\\ 
		\midrule
		\multirow{2}*{UF\cite{Zhang:08}} & UF1 & Concave  & Separable	\\ 
		~ & UF2 & Concave  & Separable\\ 
		\midrule
		\multirow{6}*{WFG\cite{Huband:06}} & WFG1 & Convex  & Separable	\\ 
		~ & WFG2 & Convex, disconnected  & Partially separable	\\ 
		~ & WFG3 & Linear, degenerate  & Partially separable	\\ 
		~ & WFG4 & Concave  & Separable	\\ 
		~ & WFG5 & Concave  & Separable	\\ 
		~ & WFG7 & Concave  & Separable	\\ 
		\bottomrule
	\end{tabular}
\end{table}
We did not apply high-dimensional WFG6, WFG8, and WFG9 as our benchmark functions because these functions are not suitable for extending to high dimensions due to high computational cost. All benchmark functions are provide by geatpy\cite{geatpy:18} and pymoo\cite{pymoo:20}. 

\subsubsection{Compareing methods and parameters}
We combine our proposal in decomposition with NSGA-II (CC-NSGA-LMM) and comparing with Random Grouping (CC-NSGA-G)\cite{Antonio:13}, Differential Grouping (CC-NSGA-DG)\cite{Sander:18}, and Monotonicity Detection (CC-NSGA-LIMD)\cite{Izumiya:17} with 30 trial runs. Besides, we compare our ultimate proposal (CC-hNSGA-LMM) with CC-NSGA-LMM to verify the effect of hybrid NSGA-II. Table \ref{tbl:2} shows the parameters of our proposal in the grouping stage, and Table \ref{tbl:3} shows the parameters of sub-problems optimization. The NSGA-II is also provided by geatpy. 
\begin{table}[tbh]
	\scriptsize
	\centering
	\caption{The parameters of decomposition optimization}
	\label{tbl:2}
	\begin{tabular}{cc}
		\toprule
		Parameter & value \\
		\midrule
		Optimization direction & Minimization	\\ 
		Optimizer & Elitist GA	\\ 
		Population size & 20	\\ 
		Max iteration& 20	\\ 
		Gene length  & 6 and 7	\\ 
		\bottomrule
	\end{tabular}
\end{table}

\begin{table}[tbh]
	\scriptsize
	\centering
	\caption{The parameters of sub-problems optimization}
	\label{tbl:3}
	\begin{tabular}{cc}
		\toprule
		Parameter & value \\
		\midrule
		Dimension & $500$-D and $1000$-D \\
		FEs & 750,000 and 1,500,000	\\ 
		Optimization direction & Minimization	\\ 
		Optimizer & NSGA-II	\\ 
		Population size & 50 \\ 
		Crossover rate & 0.9	\\ 
		Mutation rate   & 0.2	\\ 
		\bottomrule
	\end{tabular}
\end{table}

\subsubsection{Performance indicators}
To evaluate the performance of our proposal, we introduced Hypervolume (HV)\cite{Zitzler:98} and Inverted Generational Distance (IGD)\cite{Li:19} as indicators to evaluate the performance of the algorithms. HV was first presented as the size of the space covered. Given a solution set $A$ and a reference point $r$, HV can be calculated as Eq (\ref{eq:17})
\begin{equation}
	\label{eq:17}
	\begin{aligned}
		HV(A)=\lambda(\bigcup_{a \in A} \{x\vert a \prec x \prec r\})
	\end{aligned}
\end{equation}
where $\lambda$ denotes the Lebesgue measure. IGD is also the most commonly used indicator. given a solution set $A$ and a reference set $R = \{r_1, r_2, ..., r_M \}$, IGD can be defined as Eq (\ref{eq:18})
\begin{equation}
	\label{eq:18}
	\begin{aligned}
		IGD(A, R)=\frac{1}{M}(\sum_{i=1}^{M}\mathop{{\rm min}}_{a \in A} \ {\rm dis}(r_i,a))
	\end{aligned}
\end{equation}
where ${\rm dis}(r_i, a)$ denotes the Euclidean distance between $r_i$ and $a$, and a lower IGD value means better performance as same as HV.

\subsection{Performance of our proposal}
\begin{figure*}[htb]
	\centering
	\includegraphics[width=16cm]{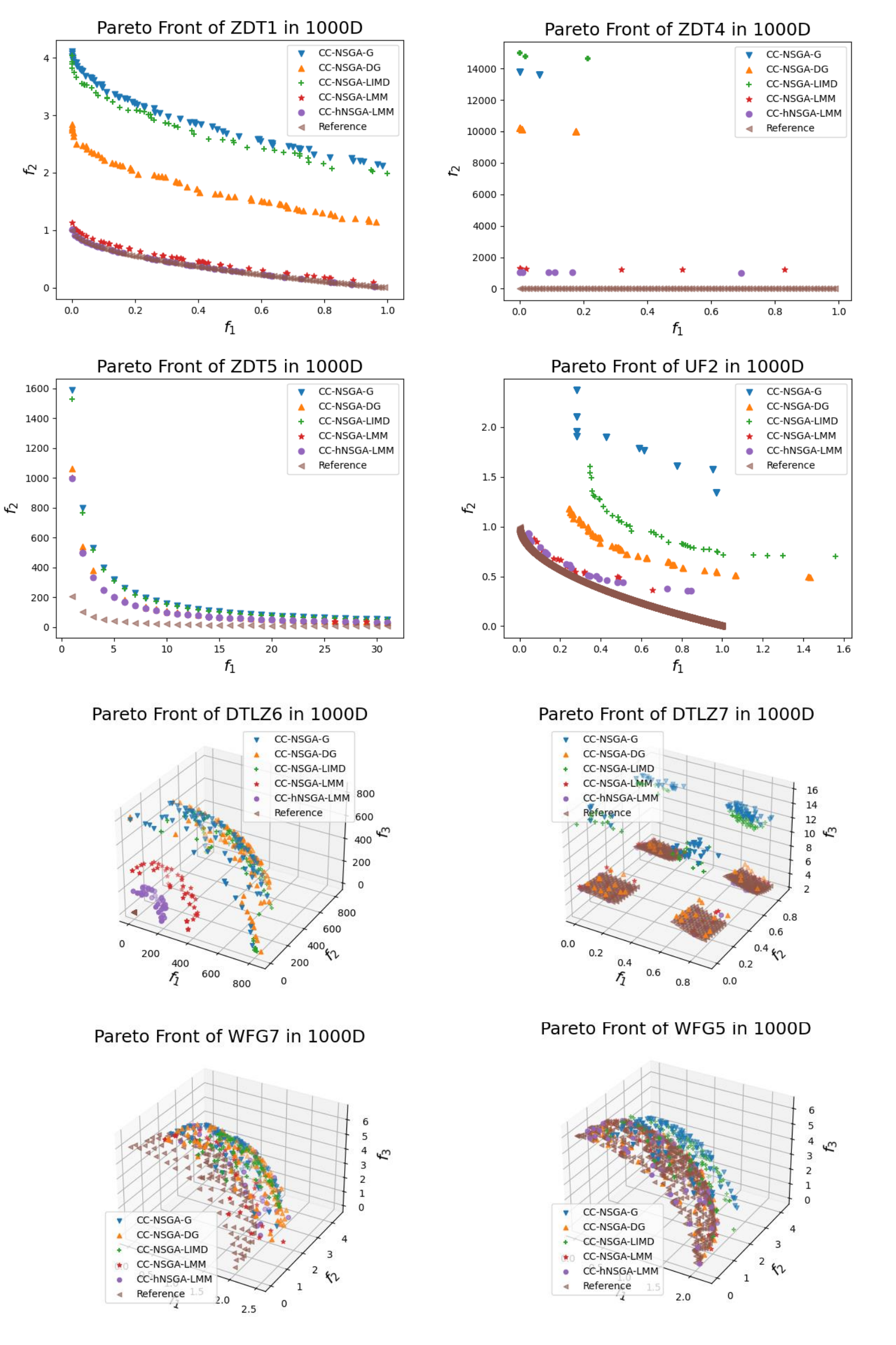}
	\caption{The representive PF graphs within 5 methods and reference sets in $500$-D and $1000$-D.}
	\label{fig:5}
\end{figure*}

In this section, the performance of CC-hNSGA-LMM is studied, both on our proposed decomposition method and the introduction of Gaussian sampling based on an estimated convergence point. Experiments are conducted on the benchmark functions presented in Section IV-A1 with 30 independent runs. Besides, we randomly choose one trial run result in 30 trial runs and draw the PF graph within comparing methods and reference sets. Due to space limitations, we select some representative PF graphs in Figure\ref{fig:5}. The mean of HV and IGD calculated in 30 trial runs are shown in Table \ref{tbl:4} and Table \ref{tbl:5}. The best solution among CC-NSGA-G, CC-NSGA-DG, CC-NSGA-LIMD, and CC-NSGA-LMM these 4 methods is highlighted with \textcolor{red}{red} and \textcolor{orange}{orange} in $500$-D and $1000$-D respectively to show the performance of our proposed decomposition method. Besides, we mark the better solution between CC-NSGA-LMM and CC-hNSGA-LMM with $^+$ to show the effect of the introduction of Gaussian sampling based on an estimated convergence point. The FEs consumed in the decomposition stage of DG, LIMD, and LMM are provided in Table \ref{tbl:6}.

\begin{figure*}[htb]
	\centering
	\includegraphics[width=16cm]{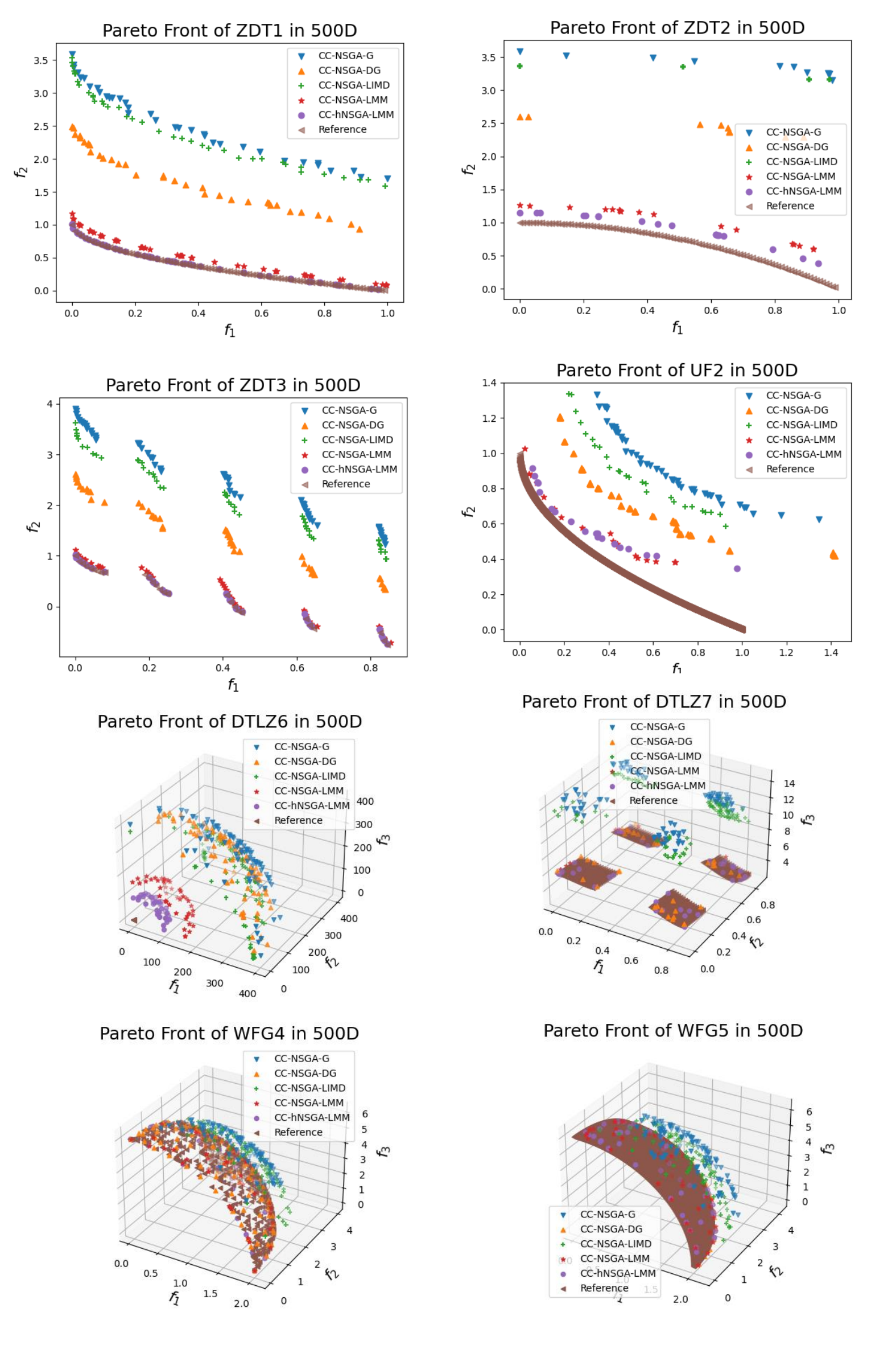}
\end{figure*}

\begin{table*}[]
	\scriptsize
	\centering
	\caption{The mean of HV among 5 methods in 30 trial runs}
	\label{tbl:4}
	\begin{tabular}{ccccccccccc}
		\toprule
		\multirow{2}{*}{Func} & \multicolumn{2}{c}{CC-NSGA-G} & \multicolumn{2}{c}{CC-NSGA-DG} & \multicolumn{2}{c}{CC-NSGA-LIMD} & \multicolumn{2}{c}{CC-NSGA-LMM} & \multicolumn{2}{c}{CC-hNSGA-LMM}  \\
		\cmidrule(r){2-3} \cmidrule(r){4-5} \cmidrule(r){6-7} \cmidrule(r){8-9} \cmidrule(r){10-11} 
		& $500$-D & $1000$-D & $500$-D & $1000$-D & $500$-D & $1000$-D & $500$-D & $1000$-D & $500$-D & $1000$-D \\
		\midrule
		ZDT1 & 0.913 & 1.147 & 0.581 & 0.710 & 0.872 & 1.101 & \textcolor{red}{0.158} & \textcolor{orange}{0.177} & 0.134$^+$ & 0.144$^+$ \\
		ZDT2 & 0.918 & 1.517 & 0.523 & 0.784 & 0.851 & 1.221 & \textcolor{red}{0.075} & \textcolor{orange}{0.296} & 0.071$^+$ & 0.148$^+$ \\
		ZDT3 & 0.499 & 1.376 & 0.193 & 0.714 & 0.443 & 0.907 & \textcolor{red}{0.177} & \textcolor{orange}{0.195} & 0.150$^+$ & 0.083$^+$ \\
		ZDT4 & 1690.193 & 1578.118 & 1006.923 & 773.871 & 1615.125 & 1360.387 & \textcolor{red}{149.293} & \textcolor{orange}{265.902} & 112.492$^+$ & 168.336$^+$ \\
		ZDT5 & 1408.709 & 2175.905 & 963.814 & 1511.787 & 1395.165 & 2058.472 & \textcolor{red}{947.672} & \textcolor{orange}{1506.865} & 885.373$^+$ & 1262.068$^+$ \\
		ZDT6 & 0.431 & 0.355 & 0.384 & 0.422 & 0.507 & 0.270 & \textcolor{red}{0.335} & \textcolor{orange}{0.221} & 0.279$^+$ & 0.192$^+$ \\
		DTLZ1 & 8.261e10 & 6.956e11 & 30228.461 & 2.403e11 & 7.210e10 & 8.988e11 & \textcolor{red}{26869.899} & \textcolor{orange}{2.111e5} & 1440.654$^+$ & 9060.585$^+$ \\
		DTLZ2 & 706.197 & 10266.957 & \textcolor{red}{0.073} & 4269.415 & 428.909 & 9674.223 & 0.077 & \textcolor{orange}{0.067} & 0.063$^+$ & 0.056$^+$ \\
		DTLZ3 & 4.158e12 & 7767.971 & 1.780e6 & 8.773e12 & 3.251e12 & 4.632e13 & \textcolor{red}{1.545e6} & \textcolor{orange}{9.983e6} & 1.271e5$^+$ & 8.328e5$^+$ \\
		DTLZ4 & 1011.450 & 14788.623 & \textcolor{red}{0.043} & 50.488 & 539.192 & 11120.653 & 0.059$^+$ & \textcolor{orange}{0.099} & 0.060 & 0.091$^+$ \\
		DTLZ5 & 538.285 & 7767.971 & \textcolor{red}{0.161} & 4270.747 & 290.810 & 6144.751 & 0.175 & \textcolor{orange}{0.251} & 0.148$^+$ & 0.211$^+$ \\
		DTLZ6 & 5.716e6 & 5.145e7 & 4.690e6 & 4.994e7 & 3.746e6 & 4.385e7 & \textcolor{red}{2.103e5} & \textcolor{orange}{1.997e6} & 4.035e4$^+$ & 3.554e5$^+$ \\
		DTLZ7 & 1.184 & 1.886 & 0.174 & 0.445 & 0.970 & 1.660 & \textcolor{red}{0.153} & \textcolor{orange}{0.364} & 0.093$^+$ & 0.317$^+$ \\
		UF1 & 1.232 & 1.580 & 0.101 & 0.388 & 0.741 & 1.274 & \textcolor{red}{0.095} & \textcolor{orange}{0.155} & 0.067$^+$ & 0.107$^+$ \\
		UF2 & 0.136 & 1.197 & 0.105 & 0.645 & 0.121 & 0.849 & \textcolor{red}{0.094} & \textcolor{orange}{0.280} & 0.089$^+$ & 0.277$^+$ \\
		WFG1 & 5.090 & 1.536 & 4.552 & 0.841 & 4.973 & 0.903 & \textcolor{red}{4.179} & \textcolor{orange}{0.884} & 3.965$^+$ & 0.520$^+$ \\
		WFG2 & 0.335 & 0.625 & 0.317 & 0.657 & \textcolor{red}{0.269} & \textcolor{orange}{0.542} & 0.309 & 0.604 & 0.289$^+$ & 0.587$^+$ \\
		WFG3 & 0.273 & 5.077 & 0.288 & 4.260 & \textcolor{red}{0.237} & 4.036 & 0.264 & \textcolor{orange}{3.897}$^+$ & 0.253$^+$ & 3.957 \\
		WFG4 & 3.698 & 4.961 & \textcolor{red}{2.855} & 3.579 & 3.592 & 3.680 & 3.032$^+$ & \textcolor{orange}{3.150} & 3.057 & 2.652$^+$ \\
		WFG5 & 0.795 & 3.219 & 0.140 & 1.783 & 0.692 & 2.617 & \textcolor{red}{0.130} & \textcolor{orange}{1.323} & 0.129$^+$ & 0.981$^+$ \\
		WFG7 & 2.785 & 6.560 & 2.233 & 6.264 & 2.733 & 6.283 & \textcolor{red}{2.136} & \textcolor{orange}{6.114} & 2.010$^+$ & 5.842$^+$ \\
		\bottomrule
	\end{tabular}
\end{table*}

\begin{table*}[]
	\scriptsize
	\centering
	\caption{The mean of IGD among 5 methods in 30 trial runs}
	\label{tbl:5}
	\begin{tabular}{ccccccccccc}
		\toprule
		\multirow{2}{*}{Func} & \multicolumn{2}{c}{CC-NSGA-G} & \multicolumn{2}{c}{CC-NSGA-DG} & \multicolumn{2}{c}{CC-NSGA-LIMD} & \multicolumn{2}{c}{CC-NSGA-LMM} & \multicolumn{2}{c}{CC-hNSGA-LMM}  \\
		\cmidrule(r){2-3} \cmidrule(r){4-5} \cmidrule(r){6-7} \cmidrule(r){8-9} \cmidrule(r){10-11} 
		& $500$-D & $1000$-D & $500$-D & $1000$-D & $500$-D & $1000$-D & $500$-D & $1000$-D & $500$-D & $1000$-D \\
		\midrule
		ZDT1 & 1.171 & 1.474 & 0.627 & 0.760 & 1.097 & 1.379 & \textcolor{red}{0.025} & \textcolor{orange}{0.016} & 0.011 $^+$ & 0.007$^+$ \\
		ZDT2 & 2.492 & 3.012 & 1.533 & 1.800 & 2.306 & 2.842 & \textcolor{red}{0.280} & \textcolor{orange}{0.273} & 0.205 $^+$ & 0.190$^+$ \\
		ZDT3 & 0.816 & 0.981 & 0.448 & 0.511 & 0.731 & 0.899 & \textcolor{red}{0.014} & \textcolor{orange}{0.015} & 0.008 $^+$ & 0.007$^+$ \\
		ZDT4 & 6106.310 & 13626.344 & 3659.787 & 8258.761 & 5861.058 & 13467.469 & \textcolor{red}{539.775} & \textcolor{orange}{1151.956} & 403.140 $^+$ & 845.605$^+$ \\
		ZDT5 & 10.899 & 10.470 & 7.350 & 8.613 & 9.116 & 10.594 & \textcolor{red}{7.183}$^+$ & \textcolor{orange}{7.153} & 7.194  & 6.221$^+$ \\
		ZDT6 & 1.224 & 1.152 & 1.182 & 1.261 & \textcolor{red}{0.976} & 0.887 & 1.053 & \textcolor{orange}{0.816} & 0.938 $^+$ & 0.784$^+$ \\
		DTLZ1 & 10183.928 & 21115.183 & 77.196 & 3432.782 & 9639.649 & 22587.874 & \textcolor{red}{76.844} & \textcolor{orange}{154.833} & 28.667 $^+$ & 53.545$^+$ \\
		DTLZ2 & 19.094 & 45.582 & \textcolor{red}{0.065} & 7.940 & 16.762 & 45.225 & 0.067 & \textcolor{orange}{0.133} & 0.006 $^+$ & 0.002$^+$ \\
		DTLZ3 & 33931.424 & 73316.697 & \textcolor{red}{307.394} & 11939.406 & 31664.480 & 75494.883 & 312.767 & \textcolor{orange}{602.455} & 131.318 $^+$ & 244.015$^+$ \\
		DTLZ4 & 21.575 & 51.164 & \textcolor{red}{0.129} & 8.361 & 17.541 & 47.096 & 0.249 & \textcolor{orange}{0.453} & 0.271 $^+$ & 0.346$^+$ \\
		DTLZ5 & 19.490 & 45.704 & 0.071 & 7.851 & 16.962 & 45.825 & \textcolor{red}{0.069} & \textcolor{orange}{0.127} & 0.002 $^+$ & 0.003$^+$ \\
		DTLZ6 & 389.946 & 785.520 & 375.773 & 807.134 & 363.994 & 792.004 & \textcolor{red}{170.141} & \textcolor{orange}{356.833} & 94.073 $^+$ & 197.733$^+$ \\
		DTLZ7 & 5.528 & 6.859 & 0.146 & 1.235 & 4.308 & 5.981 & \textcolor{red}{0.125} & \textcolor{orange}{0.133} & 0.036 $^+$ & 0.027$^+$ \\
		UF1 & 1.419 & 1.651 & 0.491 & 0.658 & 1.043 & 1.444 & \textcolor{red}{0.367}$^+$ & \textcolor{orange}{0.378} & 0.384  & 0.323$^+$ \\
		UF2 & 0.445 & 0.557 & 0.217 & 0.278 & 0.332 & 0.454 & \textcolor{red}{0.076}$^+$ & \textcolor{orange}{0.081}$^+$ & 0.082  & 0.091 \\
		WFG1 & 3.798 & 3.870 & 1.614 & 1.999 & 3.697 & 3.882 & \textcolor{red}{1.503} & \textcolor{orange}{1.502} & 1.502 $^+$ & 1.405$^+$ \\
		WFG2 & 0.719 & 0.732 & 0.709 & 0.720 & \textcolor{red}{0.590} & \textcolor{orange}{0.586} & 0.630 & 0.621 & 0.610 $^+$ & 0.612$^+$ \\
		WFG3 & 1.538 & 1.078 & 0.894 & 0.895 & \textcolor{red}{0.863} & 0.908 & 0.883 & \textcolor{orange}{0.835}$^+$ & 0.871 $^+$ & 0.871 \\
		WFG4 & 0.561 & 0.496 & \textcolor{red}{0.009} & 0.089 & 0.739 & 0.747 & 0.012$^+$ & \textcolor{orange}{0.015} & 0.015 & 0.009$^+$ \\
		WFG5 & 3.213 & 0.720 & 3.020 & 0.228 & 3.252 & 0.712 & \textcolor{red}{2.989} & \textcolor{orange}{0.149} & 2.835 $^+$ & 0.137$^+$ \\
		WFG7 & 0.656 & 0.714 & 0.590 & 0.649 & 0.639 & 0.676 & \textcolor{red}{0.587} & \textcolor{orange}{0.595}$^+$ & 0.568 $^+$ & 0.605 \\
		\bottomrule
	\end{tabular}
\end{table*}

\begin{table*}[]
	\scriptsize
	\centering
	\caption{FEs consumed on decomposition stage of three variable grouping methods}
	\label{tbl:6}
	\begin{tabular}{ccccccc}
		\toprule
		\multirow{2}{*}{Func}  & \multicolumn{2}{c}{DG} & \multicolumn{2}{c}{LIMD} & \multicolumn{2}{c}{LMM}  \\
		\cmidrule(r){2-3} \cmidrule(r){4-5} \cmidrule(r){6-7} 
		& $500$-D & $1000$-D & $500$-D & $1000$-D & $500$-D & $1000$-D  \\
		\midrule
		ZDT1 & 63250 & 251500 & 1974 & 3972 & 1503 & 3003\\
		ZDT2 & 63250 & 251500 & 1974 & 3972 & 1503 & 3003\\
		ZDT3 & 63250 & 251500 & 1974 & 3972 & 1503 & 3003\\
		ZDT4 & 63250 & 251500 & 1974 & 3972 & 1503 & 3003\\
		ZDT5 & 250500 & 1001000 & 1974 & 3972 & 1503 & 3003\\
		ZDT6 & 244558 & 989058 & 1974 & 3972 & 1503 & 3003\\
		DTLZ1 & 249502 & 999002 & 1974 & 3972 & 74649 & 149304\\
		DTLZ2 & 249502 & 999002 & 1974 & 3972 & 1503 & 3003\\
		DTLZ3 & 249502 & 999002 & 1974 & 3972 & 74652 & 149274\\
		DTLZ4 & 249502 & 999002 & 1974 & 3972 & 1503 & 3003\\
		DTLZ5 & 249502 & 999002 & 1974 & 3972 & 1503 & 3003\\
		DTLZ6 & 1976 & 3974 & 1974 & 3972 & 74629 & 149271\\
		DTLZ7 & 250500 & 1001000 & 1974 & 3972 & 1503 & 3003\\
		UF1 & 250500 & 1001000 & 1974 & 3972 & 1503 & 3003\\
		UF2 & 250500 & 1001000 & 1974 & 3972 & 1503 & 3003 \\
		WFG1 & 249502 & 999002 & 1974 & 3972 & 1503 & 3003\\
		WFG2 & 247518 & 999002 & 63250 & 251500 & 74364 & 149325\\
		WFG3 & 247518 & 999002 & 63250 & 251500 & 73253 & 148245\\
		WFG4 & 249502 & 999002 & 1974 & 3972 & 1503 & 3003\\
		WFG5 & 249502 & 999002 & 63250 & 251500 & 1503 & 3003\\
		WFG7 & 248506 & 997006 & 1974 & 3972 & 1503 & 3003\\
		\bottomrule
	\end{tabular}
\end{table*}

\subsection{Analysis}
In this Section, we will analyze the effect of the LMM in decomposition and the Gaussian sampling in optimization.

\subsubsection{LMM in decomposition}
From Table \ref{tbl:4} and Table \ref{tbl:5}, we can see our proposed CC-NSGA-LMM outperforms the compared three methods in the majority of test functions. This is mainly due to the following aspects. (1). We calculate the linkage measurement function based on multiple samples. Although this is a necessary condition for the employment of LINC-R and LIMD in low-dimensional space, the linkage is only possible to check the nonlinearity around a sample point in high-dimensional space due to the FEs limitation, such as DG. Therefore, our proposal is more robust for solving large-scale optimization problems based on CC. (2). Although our proposal is based on LINC-R, the existence of the penalty allows our proposal to ignore some weak interactions between variables. This process will increase the error in the optimization stage, it can accelerate the sub-problems optimization, especially under the limitation of FEs.

From Table \ref{tbl:6}, we notice that in our proposed decomposition method, $1,503$ and $3,003$ evaluation times appear in the $500$-D and $1000$-D test functions frequently, This is because our proposal identifies these test functions as fully separable, which is consistent with the description of functions in Table \ref{tbl:1}. According to the FEs consumed by DG in the decomposition stage, it can be seen that when FEs are approximately equal to $250,000$ and $1,000,000$ in $500$-D and $1000$-D functions respectively, DG identifies the problem as a separable function. And the FEs saved in the decomposition stage allow more FEs to be allocated to optimize the sub-problems, which makes our proposal better than the compared methods.

Meanwhile, we notice that CC-NSGA-DG and CC-NSGA-LIMD outperform our proposal in some $500$-D test functions, such as DTLZ2-5. But this competitiveness almost disappeared in the $1000$-D test function. This is because the time complexity of DG and LIMD is $O(MN^2)$. $M$ is the population size, and $N$ is the dimension. DG sets $M=1$ in high-dimensional space. So as the dimension increases, more FEs are necessary to identify the interactions between variables. This is the main reason for the rapid degeneration of DG and LIMD when the dimension approaches to $1000$-D. And our proposal can control the time complexity by controlling some hyperparameters, which enables our proposal to have more feasibility to extend to larger-scale optimization problems.

However, we notice that in ZDT1-4, DG identifies these functions as partially separable functions, and DG identifies DTLZ6 as completely non-separable functions, which is inconsistent with the separability description in Table \ref{tbl:1}. Let's take ZDT1 as an example. The formula of ZDT1 is shown in Eq (\ref{eq:23})
\begin{equation}
	\label{eq:23}
	\left\{
	\begin{aligned}
		\min f_1(x_1)=x_1 \\
		\min f_2(x)=g(1-\sqrt{f_1/g}) \\
		g(x)=1+9\sum_{i=2}^{m}x_i /(m-1) \\
		s.t. \ 0 \leq x_i \leq 1 \\
	\end{aligned}
	\right.
\end{equation}
We can see that although $f_2$ in Eq (\ref{eq:23}) is a separable function at the monotonicity level in the limited search space, DG cannot detect this information, which reveals the limitation of DG in identifying the separability of such variables. At the same time, our proposal LMM identifies ZDT1-4 as fully separable functions. Due to the existence of the penalty, which weakens the effect of the fitness difference term in the linkage measurement function, so the penalty term dominates the direction of optimization.

Meanwhile, LMM recognizes DTLZ1, DTLZ3, and DTL6 as partially separable functions, which is because the fitness difference term still occupies a large proportion in the linkage measurement function. In future research, we can adaptively adjust the intensity of the penalty through some methods such as machine learning and reinforcement learning.

\subsubsection{Gaussian sampling based on an estimated convergence point}
From Table \ref{tbl:4} and Table \ref{tbl:5}, we can say the introduction of the Gaussian sampling based on an estimated convergence point can accelerate the convergence of optimization and find better PFs, which proves the hypothesis we proposed in Section I is true. Paper\cite{Yan:19} has already proved that only an estimated convergence point may not accelerate the convergence with numerical experiments, especially in multimodal functions. However, our proposal does not rely on the estimated convergence point in excess but considers the area centered on the estimated convergence point as the trust region. This strategy gives more possibilities for exploitation, although it needs to consume some FEs.

\section{Discuss}
The above analysis shows our proposal has broad prospects to solve LSMOPs, however, there are still many aspects for improvement. Here, we list a few open topics for potential and future research.

\subsection{The design of linkage measurement function}
In our proposed variable grouping method, we design a linkage measurement function to lead the direction of decomposition solution search, and there are mainly three important constituent elements of this function: fitness difference term, penalty, and weights of the objective function. To simplify the design of the linkage measurement function, we apply constant penalty and averaging weights. In practice, the intensity of the penalty and the weights of objective functions should be changed based on the feature of LSMOPs. For example, in fully separable LSMOPs, the penalty will play a decisive role to lead the direction of optimization due to the fitness difference term being infinitely close to 0, and the intensity of the penalty will decrease with the increase of interactions between variables. And for weights of objective functions, the importance of different objective functions should be different in practice. Adaptively deciding these hyperparameters based on reinforcement learning or machine learning is our future research topic. 

\subsection{More powerful local search operator}
In this experiment, we design a Gaussian sampling operator with the mean of an estimated convergence point and a constant variance as our local search operator and achieved good experimental results. The search space of the optimization problem is often different, and the search space corresponding to the concept of "local" is also different. Given a simple example, the search space of two problems are $[-0.5, 0.5]$ and $[-500, 500]$ respectively, so it is unreasonable to apply a constant as the Gaussian sampling variance for all kinds of optimization problems. Therefore, an adaptive variance determination strategy is necessary. In future research, it is promising research to introduce the CMA-ES\cite{Hansen:16} as our local search operator.

\subsection{The scalability of our proposal}
This paper mainly contributes two different aspects to accelerate the convergence of LSMOPs. (1). Designing a novel decomposition method. (2). Developing a efficient Gaussian samlping operator based on the estimated convergence point combined with NSGA-II. 

For decomposition method, our proposal performed well on $500$-D and $1000$-D LSMOPs. In the face of higher dimensional problems, The computational cost of DG, especially on fully separable functions, is completely unacceptable\cite{Omidvar:14}. However, our proposal can control the consumption of FEs by adjusting some hyperparameters, such as the length of genes (number of groups), the maximum iteration of decoposition optimization, etc. And the previous research shows that the proposed variable grouping method can identify the interactions between variables in noisy environments. Therefore, our proposed grouping method has stronger scalability than the current popular variable grouping methods theoretically. In the future, the introduction of our proposal to solve higher-dimensional LSMOPs, constraint LSMOPs, and real-world LSMOPs is our research direction.

For the Gaussian sampling operator, experimental results show that our proposal can accelerate the convergence and find better PF in solving LSMOPs. In theory, our designed Gaussian sampling operator has strong scalability to combine with GA, DE, and PSO in single-objective evolutionary algorithms (SOEAs), and MOEA/D, NSGA in MOEAs with simple modification. In the future, the combination of our proposal with novel EAs to solve more complex optimization problems is a promising research topics.

\section{Conclusion}
In this paper, we extend our previous research on the variable grouping method for LSSOPs to LSMOPs and design a Gaussian sampling operator based on an estimated convergence point to accelerate the convergence of the optimizer to find PF. To evaluate our proposal, we conduct our experiments on $500$-D and $1000$-D test functions and compare our proposal with popular methods. Experiments show that our proposed variable grouping method is better than the three compared methods, and our proposed hNSGA can significantly accelerate the convergence in optimization. At the end of this paper, we list some interesting topics which can improve our algorithm. Finally, our proposal has broad prospects for addressing LSMOPs.

\section{Acknowledgement}
This work was supported by JSPS KAKENHI Grant Number JP20K11967.

\bibliographystyle{IEEEtran}
\bibliography{Paper}

\end{document}